\documentclass[sigconf]{acmart}
\AtBeginDocument{%
  \providecommand\BibTeX{{%
    \normalfont B\kern-0.5em{\scshape i\kern-0.25em b}\kern-0.8em\TeX}}}

\usepackage{times}
\usepackage{epsfig}
\usepackage{graphicx}
\usepackage{amsmath}
\usepackage{bbding}
\usepackage[tone,extra,safe]{tipa}

\usepackage[capitalize]{cleveref}
\usepackage{expl3}
\ExplSyntaxOn
\newcommand\latinabbrev[1]{
	\peek_meaning:NTF . {
		#1\@}%
	{ \peek_catcode:NTF a {
			#1.\@ }%
		{#1.\@}}} 
\ExplSyntaxOff

\newcommand{\para}[1]{{\vspace{0.2cm}\noindent\textbf{#1}}}

\def\eg{\latinabbrev{e.g}}

\def\ie{\latinabbrev{i.e}}

\DeclareMathOperator*{\argmin}{arg\,min}

\acmSubmissionID{373}
\setcopyright{acmcopyright}
\copyrightyear{2023}
\acmYear{2023}

\begin{document}

\title{Rethinking Voice-Face Correlation: A Geometry View}

\author{Xiang Li$^1$, Yandong Wen$^2$, Muqiao Yang$^1$, Jinglu Wang$^3$, Rita Singh$^1$, Bhiksha Raj$^{1,4}$}
\affiliation{
$^1$ Carnegie Mellon University, $^2$ Max Planck Institute, $^3$ Microsoft,\\ $^4$ Mohamed bin Zayed University of Artificial Intelligence\city{}\country{}}



\begin{abstract}
Previous works on voice-face matching and voice-guided face synthesis demonstrate strong correlations between voice and face, but mainly rely on coarse semantic cues such as gender, age, and emotion. In this paper, we aim to investigate the capability of reconstructing the 3D facial shape from voice from a geometry perspective without any semantic information. We propose a voice-anthropometric measurement (AM)-face paradigm, which identifies predictable facial AMs from the voice and uses them to guide 3D face reconstruction. By leveraging AMs as a proxy to link the voice and face geometry, we can eliminate the influence of unpredictable AMs and make the face geometry tractable. Our approach is evaluated on our proposed dataset with ground-truth 3D face scans and corresponding voice recordings, and we find significant correlations between voice and specific parts of the face geometry, such as the nasal cavity and cranium. Our work offers a new perspective on voice-face correlation and can serve as a good empirical study for anthropometry science. 
\end{abstract}
\maketitle
\section{Introduction}
The study of face-voice correlation has been extensively investigated in recent years. Previous works on voice-face matching \cite{Wen_2021_CVPR,ning2021disentangled,zheng2021adversarial}, voice-guided face synthesis \cite{jamaludin2019you, zhou2019talking, guo2021adnerf,cudeiro2019capture}, and voice-guided face modification have indicated a strong correlation between voice and face. The most intuitive and commonly used consensus encoded between voice and face is mainly based on semantics, such as gender, age and emotion. Most prior works aim to learn a semantic correspondence between voice and face and conduct crossmodal tasks by leveraging those consensuses. For example, for voice-guided face synthesis, the generated faces have reasonable appearances with proper gender, age and emotion status corresponding to the voice. Those semantic correlations are strong and easy to learn thus dominant previous models while a fundamental question we want to cast is, are there any other voice-face correlations except for those coarse semantics? Is reconstructing identity-fidelity 3D face from voice possible?
In this paper, we aim to explore the voice-face correlations in a geometry view after constraining all those easily learned semantic biases.

\begin{figure}[t!]
    \centering    \includegraphics[width=\linewidth]{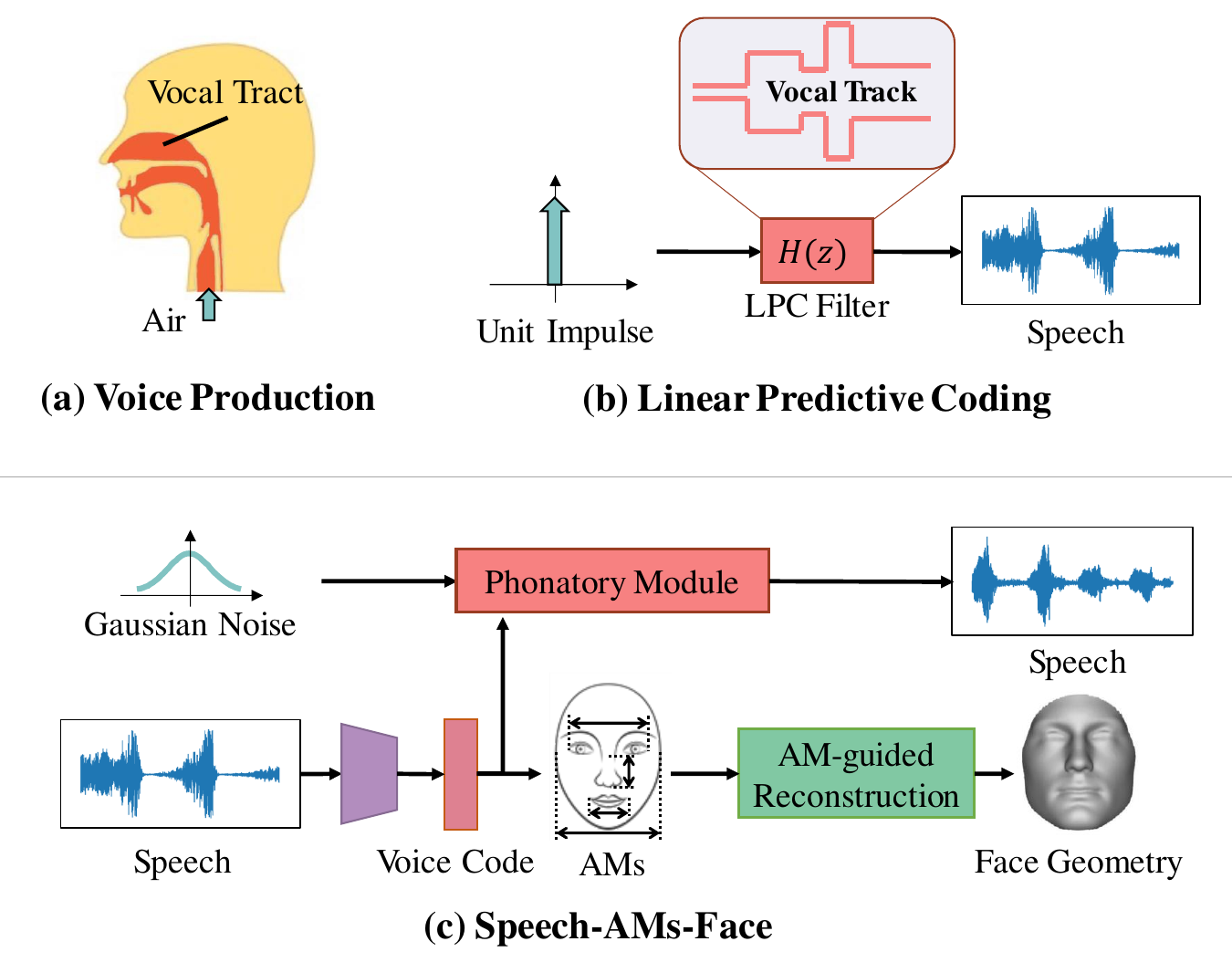}
    \vspace{-0.8cm}
    \caption{(a) Human voice production. (b) Linear predictive coding represents the voice by a unit impulse with a set of linear filters which can be interpreted as an estimation of the vocal tract. (c) Our voice-AM-Face pipeline first predicts and verifies predictable anthropometric measurements (AMs) and then utilizes AMs to guide 3D face reconstruction. A phonatory module is involved to obtain a better representation for AM prediction.}
    \label{fig:teaser}
\end{figure}

There are several previous works investigating recovering face from voice. Most of them are from a 2D perspective \cite{jamaludin2019you, zhou2019talking, guo2021adnerf}, which utilize Generative Adversarial Network (GAN) \cite{mirza2014conditional,goodfellow2020generative} to generate faces with voice as the condition. However, face recovering from voice is ill-posed. \cite{oh2019speech2face} found that the recovery mainly focuses on some semantics of the speaker. For example, attributes such as ethnicity has weak or no function while gender and age tend to be recovered. Since those models mainly rely on semantics, the results are not identity-fidelity which means generated faces can look very different from the original ones. In addition, for a 2D face image, identity-unrelated factors like expressions, hairs, glasses, illumination, background, etc., are also involved in the recovery process leading to noisy and unstable outcomes. Different from 2D images, general 3D facial shape is represented by the 3D coordinates of a number of points on its surface called vertices \cite{blanz1999morphable} which inherently excludes the identity-unrelated factors. Moreover, since the topology of 3D facial shape is predefined and consistent across different faces, we can easily measure the reconstruction accuracy with distances between the predicted vertices and their ground truths.

Similar to our target, one recent work \cite{wu2022cross} attempts to recover 3D faces from voice while, due to the lack of ground-truth 3D face scans, they first generate 2D face images from voice and then reconstruct 3D faces guided by an off-the-shelf 3D face reconstruction model. The noise enrolled in the 2D-to-3D face reconstruction makes the result unconvincing. For example, any expression in the 2D face from the first stage will force the reconstructed 3D face to have the same expression. In this way, we consider the face is still determined by the first-stage 2D face image. 

In our method, we aim to disable all previously used semantics, \eg, gender, age and emotion, and focus on the voice-face correlation from a pure geometry view. Before introducing our method, let us go back and understand how voice is generated by human beings. voice is produced by phonatory structures (\cref{fig:teaser} (a)), \eg, vocal tract and vocal cords. Specifically, when producing vowels, the vocal cords vibrate with no obstruction in the vocal tract. In contrast, for most consonants, the phonation purely depends on the vocal tract resonance with a pulmonic airflow. The vocal tract can be assumed as a filter that makes the phonemes versatile and personalized. With the phonation mechanism of human beings, as shown in \cref{fig:teaser} (b), Markel \textit{et al} introduces linear predictive coding (LPC) \cite{markel1976linear} which models phonation as a unit impulse signal modified by a stack of tubes (vocal tract) and encodes personalized voice by vocal tract coefficients. The LPC yields a good physical model of the vocal tract with only voice inputs in an unsupervised manner. As the mouth and nose serve as the most important parts of the vocal tract, we hypothesize that their geometry should be encoded in the voice. With the tight bind of muscles and skeletons, other parts of face geometry may also be represented by voice.

Though voice and face geometry should have some correlations, we have no idea about which part of the face voice can represent. Constructing uncorrelated relations will lead to random results and raise the model instability. To tackle this problem, we introduce the voice-anthropometric measurement (AM)-face paradigm. Previous studies have shown that anthropometric measurements like the dimensions of nasal cavities \cite{vampola2020influence} or cranium \cite{wyganowska2013anthropometric,wyganowska2017vocal} directly influence the speaker's voices. In our voice-AM-face paradigm, we first summarize a set of AMs from anthropometry literature \cite{ramanathan2006modeling,zhuang2010facial,shan2021anthropometric,farkas2004anthropometric,ghafourzadeh2019part}, then identify predictable AMs and use them to guide the 3D face reconstruction by conducting AM-guided optimization. By leveraging AMs as a proxy to link the voice and face geometry, we can eliminate the influence of unpredictable AMs and make the face geometry tractable. In addition, the analysis of AMs also brings a new view to understanding voice-face correlation in a fine-grained fashion.

Inspired by LPC which learns the shape of the vocal tract by producing voice, we utilize a phonatory module to facilitate voice representation learning for face geometry. Similar to the auto-regressive impulse-by-filter model used in LPC, recently introduced denoising diffusion probabilistic models \cite{ho2020denoising} share a similar structure, which samples a random noise with auto-regressive updating to form the final result. Based on the structure similarity, we choose the diffusion model as our phonatory module. 

With the predicted AMs, we reconstruct the facial shapes by an optimization-based method, which first projects the 3D facial shapes into a low-dimensional linear space \cite{blanz1999morphable}. By adjusting the coefficients in low-dimensional space, we obtain different re-projected 3D facial shapes. Though this paper mainly focuses on understanding the relationship between the 3D facial shape and voice from a scientific angle, this technique has its potential applications. For example, the identity-fidelity facial shape can be used for criminal profiling scenarios, such as hoax calls and voice-based phishing.

In this paper, we try to answer two core questions - (1) Is there a correlation between face geometry and voice?  (2) If so, which part of the face can be represented by the voice? To fulfill our target, we collect a large-scale dataset containing ground-truth 3D face scans and corresponding voice recordings from 1026 speaker identities. A voice-AM-face paradigm equipped with a phonatory module is proposed for analyzing the voice-face correlation. Our contributions can be summarized as follows.
\begin{itemize}
    \item We propose a voice-AM-face paradigm and a corresponding voice-face dataset for tractable 3D face deduction from voice. 
    \item We investigate voice-face correlation in a fine-grained manner by statistically verifying which part of the face can be reflected by the voice. The results can serve as a good reference to support future voice-face research, such as voice-face verification.
    \item We leverage voice production as a proxy task to learn face geometry representation and verify that voice production is highly related to 3D facial shapes.
\end{itemize}


\section{Related Works}

\subsection{Voice-face Matching and Voice-guided Face Synthesis}

The human voice contains rich information that can be used to recognize personality traits, such as speaker identity \cite{bull1983voice, maguinness2018understanding, ravanelli2018speaker}, gender \cite{li2019improving}, age \cite{ptacek1966age, singh2016relationship, grzybowska2016speaker}, and emotion status \cite{wang2017learning, zhang2019attention}. Voices can also be used for monitoring health conditions \cite{ali2017automatic} and other medical applications \cite{han2021exploring}. Most existing works in this area focus on predicting personality traits that are intuitively related to voice. Such personality traits may have essential correlations between the human voice and their faces \cite{NEURIPS2019_eb9fc349}.

Cross-modal voice-face matching \cite{Wen_2021_CVPR,ning2021disentangled,zheng2021adversarial} and cross-modal verification \cite{Nawaz_2021_CVPR, tao2020audio,sari2021multi} are tasks where voices are used as queries to retrieve faces or vice versa, which have received increasing attention in recent years. Voice-guided face synthesis is another related task, which aims to generate coherent and natural lip movements, and includes methods that drive template images \cite{jamaludin2019you, zhou2019talking, guo2021adnerf} or template face meshes \cite{cudeiro2019capture} to talk by speech inputs, or replace lip movements in a video with movements inferred from another video or speech \cite{chen2018lip, wiles2018x2face}. 

Unlike the existing work in related fields that are more focused on semantic correlations between voice and face, our work investigates the voice-face correlation from a geometry view by studying holistic facial structures. There has been recent work that seeks to understand the correlations between voice and facial geometry by first recovering 2D faces from voice and then reconstructing 3D faces from the 2D representations \cite{wu2022cross}. However, during this process, it is still inevitable that the semantic correlations are encoded in the 2D face and affect the 2D-to-3D face reconstruction. Instead, we aim to model our voice-face correlation from a pure geometry view without the influence of any semantics.

\begin{figure*}[t]
    \vspace{0.2cm}
    \centering    \includegraphics[width=\linewidth]{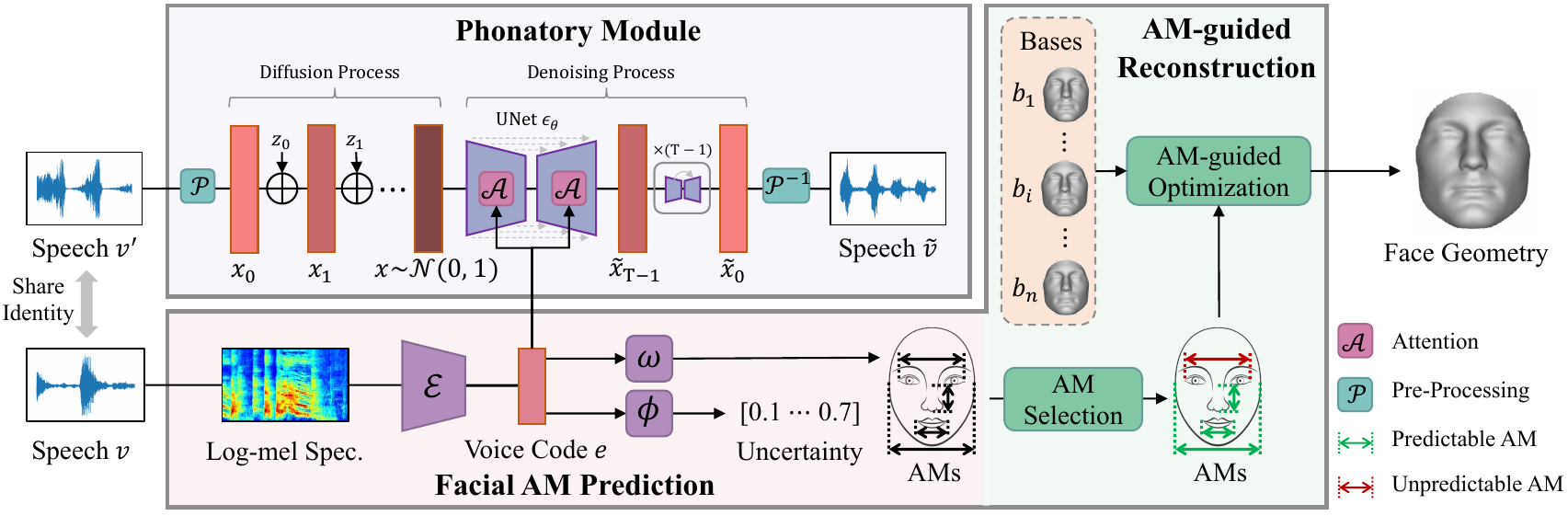}
    \caption{Illustration of our analysis pipeline for voice-face correlation. We randomly pick two voice recordings with shared speaker identity as $v$ and $v^\prime$. We then analyze the relationship between each AM and voice by predicting each AM from voice with an estimator and an intermediate voice code $e$. The optional phonatory module equips a diffusion-based voice generation model with a voice code $e$ as a condition to conduct voice style cloning to help us understand the relationship between face geometry and voice characteristics, which serves as an additional constraint to enforce the estimator learn voice identity. We analyze and select AMs with hypothesis testing. The statically significantly predictable AMs are utilized for 3D facial shape reconstruction for further analysis.}
    \label{fig:pipeline}
\end{figure*}

\subsection{Phonation and Anthropometry}

The human voice is generated by phonatory structures, and the phonation of different phonemes may be dependent on different physiological structures. For example, the phonation of consonants includes some airflow obstruction in the vocal tract, while vowels do not. By utilizing such properties, it has been proven to be informative and helpful in various tasks, including automatic speech recognition \cite{ghosh2011automatic}, speech enhancement \cite{yang2023paaploss} and emotion recognition \cite{gomez2007relationships}. Beyond those language-related usages, human attributes are also predictable from voice. There is a substantial body of research on inferring human attributes from a person's voice, including speaker identity \cite{Bull1983TheVA, Ravanelli2018SpeakerRF}, age \cite{Bahari2012AgeEF, Ptacek1966AgeRF}, gender \cite{Li2019ImprovingTS}, and emotion status \cite{wang2017learning, Zhang2019AttentionaugmentedEM}. The interaction between these physiological structures may play an important role in the recovery of 3D faces from voice. More specifically, the underlying skeletal and articulatory structure of the face and the tissue covering them may govern the shapes, sizes, and acoustic properties of the vocal tract that produces the voice. Linear predictive coding (LPC) \cite{markel1976linear} which models phonation as a unit impulse signal modified by a stack of tubes (vocal tract) and encodes personalized voice by vocal tract coefficients. The LPC yields a good physical model of the vocal tract with only voice inputs in an unsupervised manner.

To explicitly describe the correspondence between vocal and facial features, anthropometric measurements have been used in a wide range of applications to associate with voice production \cite{singh2016forensic, ramanathan2006modeling,zhuang2010facial,shan2021anthropometric,farkas2004anthropometric,ghafourzadeh2019part}. In a broad sense, AMs may cover various body parameters and characteristics, including skeletal proportions, race, height, body size, etc. These characteristics may influence the phonation of voice by the differences in the placement of the glottis, length of vocal cords, etc.

In this work, we summarize a large set of AMs that is highly associated with voice-face correlation. Meanwhile, we also identify the predictable AMs to guide the 3D facial shape reconstruction. The results can serve as a good reference to support future voice-face research.

\section{Method}
In this section, we first introduce the task formulation and then demonstrate our method in detail.
\subsection{Formulation}
We aim to reconstruct any speaker's 3D facial shape from their voice recordings. Given a set of paired voice recordings and 3D facial shapes $\{(v_i, f_i)\}$ from different individuals, where $v_i$ is a voice recording spoken by the $i$-th person and $f_i$ is a 3D facial shape scanned from the speaker of $v_i$. The goal is to reconstruct the 3D facial shape $f$ of any speaker from their voice recording $v$. In our method, we introduce anthropometric measurements (AMs) $m=\{m^{(1)},\cdots,m^{(k)}\}$ computed from $f$ as a proxy, where $K$ is a positive integer and $m^k$ ($k\in[1,K]$) denotes the $k$-th AM. Accordingly, the overall dataset is denoted as $\mathcal{D}=\{(v_i,f_i,m_i)\}$. To statistically analyze the results, we construct an additional validation set for empirically validating the dependency. Specifically, the dataset $\mathcal{D}$ is split into a training set $\mathcal{D}_t$ for model learning, a validation set $\mathcal{D}_{v1}$ for model selection, a validation set $\mathcal{D}_{v2}$ for AM selection, and an evaluation set $\mathcal{D}_e$ for evaluating the reconstructed 3D facial shapes. All splits have no overlap.

\subsection{Pipeline Overview}
As shown in \cref{fig:pipeline}, the proposed method has three main components - facial AM prediction, AM-guided reconstruction and an auxiliary phonatory module. On one hand, we predict the AMs that are potentially correlated with voice production from anthropometry literature \cite{ramanathan2006modeling,zhuang2010facial,shan2021anthropometric,farkas2004anthropometric,ghafourzadeh2019part}. An estimator $\mathcal{E}$ is trained with uncertainty learning with a voice code $e$. On the other hand, inspired by the voice production mechanism, we introduce a phonatory module as a constraint to facilitate the training of AM prediction. In particular, a diffusion-based voice generation module is involved as the phonatory module which aims to imitate the voice identity conditioning on the voice code $e$. After that, we select the AMs predictable from voice for hypothesis testing. The null hypothesis is made for each AM and states the AM is unpredictable from voice. We can successfully reject the corresponding null hypothesis if any AM estimation is better than chance on a held-out validation set with statistical significance. The final 3D facial shapes can be reconstructed by a fitting process \cite{blanz2003face} based on the predictable AMs. This is conducted by adjusting a set of coefficients in low-dimensional space, such that the differences between the AMs of the generated 3D facial shape and the predicted AMs are minimized. Intuitively, if there are more predictable AMs spanning different locations of a face, the reconstruction can be more indistinguishable.

\subsection{Facial AM Prediction}
In this section, we illustrate our method to predict facial AMs from voice.

\begin{figure}[t]
    \centering    \includegraphics[width=0.95\linewidth]{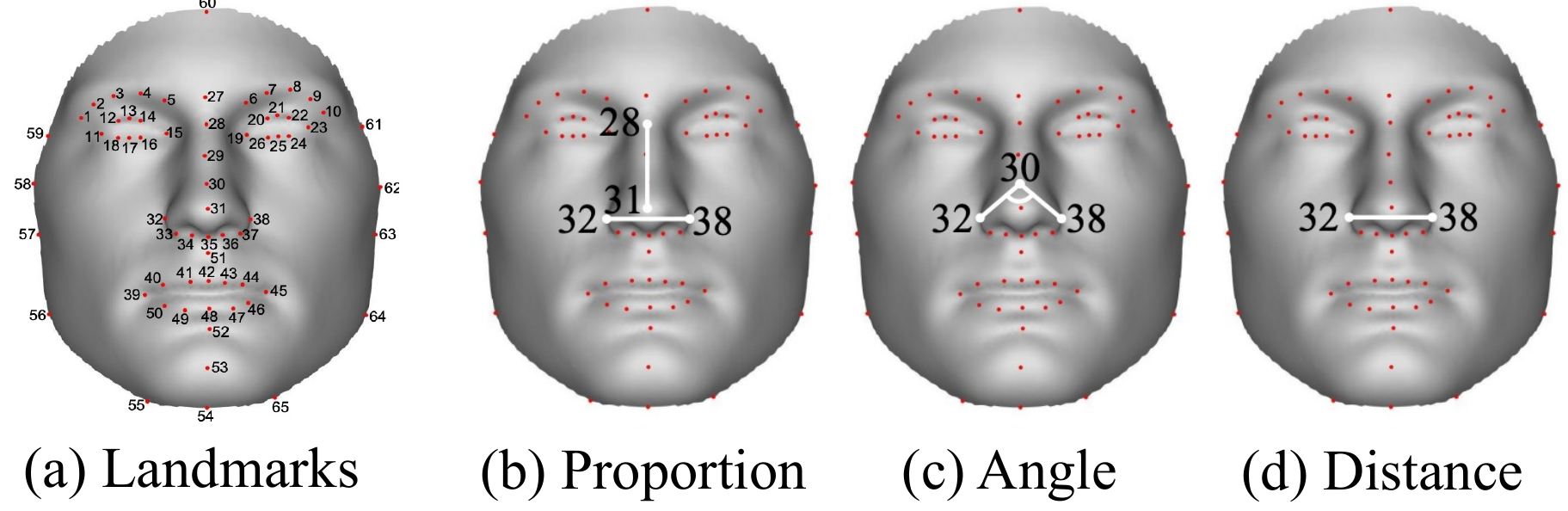}
    \caption{Examples of summarized AMs. We summarized three types of AM: proportion, angle and distance. Those AMs are computed from the predefined landmark on the 3D face representation.}
    \label{fig:AMs}
\end{figure}

\para{AM summarization.}
There is a large body of literature on anthropometry. Extensive studies show that many AMs of human faces can be associated with voice production \cite{ramanathan2006modeling,zhuang2010facial,shan2021anthropometric,farkas2004anthropometric,ghafourzadeh2019part}. We summarize the most commonly used AMs as shown in \cref{fig:AMs} (the complete list of AMs is available in the appendix). The chosen AMs are categorized as proportion, angles and distance of a set of face landmarks. Those intra-face features are more robust than 3D coordinate representations as the variations resulting from spatial misalignment are completely eliminated. 

\para{Uncertainty-aware AM estimation.}
The AM prediction is conducted by an estimator trained with an uncertainty-aware scheme. Let $F_k(v;\mathcal{E}_k, \omega_k):v\mapsto \mathbb{R}$ be an estimator that maps voice recording $v$ into the $k$-th predicted AMs, where $\mathcal{E}_k$ and $\omega_k$ are the learnable parameters. As this is a regression problem, we leverage 
\begin{equation}
    \{\mathcal{E}_k^*, \omega_k^*\}=\argmin_{\mathcal{E}_k,\omega_k}\frac{1}{|\mathcal{D}_t|}\sum_{(v,m^{(k)})\in\mathcal{D}_t}(F_k(v;\mathcal{E}_k, \omega_k)-m^{(k)})^2
\end{equation}
as the training objective for the $k$-th AM. $|\mathcal{D}_t|$ is the number of the triplets (voice, face and AMs) in dataset $\mathcal{D}_t$. By incorporating uncertainty into the estimator learning, the prediction becomes a random variable rather than a single value. We leverage a Gaussian distribution to the prediction. The estimator $F_k(v;\mathcal{E}_k,\omega_k)$ maps $v$ into the mean of the $i$-th predicted AM. Similarly, we define an uncertainty estimator $G_L(v;\mathcal{E}_k,\phi_k):v\mapsto\mathbb{R}^+\cup\{0\}$ that $v$ into the variance of the $k$-th predicted AM. Again, $\mathcal{E}_k$ and $\phi_k$ are the learnable parameters. The predicted AM and its ground truth become $\mathcal{N}(F_k(v), G_k(v))$ and $\mathcal{N}(m^{(0)},0)$ respectively \cite{kendall2017uncertainties}. Given two random variables, a more reasonable learning objective is to minimize their KL divergence.
\begin{equation}
\begin{aligned}
    \{\mathcal{E}_k^*,\omega_k^*,\phi_k^*\}=\argmin_{\mathcal{E}_k,\omega_k,\phi_k}\frac{1}{|\mathcal{D}_t|}\sum_{(v,m^{(k)})\in\mathcal{D}_t}\frac{(F_k(v;\mathcal{E}_k,\omega_k)-m^{(k)})^2}{G_k(v;\mathcal{E}_k,\phi_k)}\\+\mathrm{ln}G_k(v;\mathcal{E}_k,\phi_k)
\end{aligned}
\end{equation}
For a fixed $(F_k(v;\mathcal{E}_k,\omega_k)-m^{(k)})^2$, there is an optimal variance $G_k(v;\mathcal{E}_k,\phi_k)=(F_k(v;\mathcal{E}_k,\omega_k)-m^{(k)})^2$ such that the loss function is minimized. Thereby the uncertainty estimator $G_k$ is learned to produce a small variance if the prediction error is small and vice versa. On the contrary, a smaller variance indicates that the predicted AM is more likely to yield a small prediction error,\ie, close to the ground truth. In this way, we can choose to trust the predicted AMs when the predicted variances are small, and defer the voice recordings to human experts otherwise. An extreme case is $G_k(v)\equiv1$ where the uncertainty learning objective degrades to the regular regression model.

\para{Temporal aggregation.}
In practice, following the convention of voice understanding, the long voice recording $v$ is fed into the network in the form of multiple short segments $\{v^{(1)},\cdots,v^{(L)}\}$. We obtain a sequence of means and variances of the predicted AM. During training, we compute the loss for each segment individually and average them as the training loss. While during evaluation, the predicted AM and its uncertainty are given by aggregating the predictions among all segments. Assuming the short segments from a long recording are class-conditionally independent, the formulations of aggregation are 
\begin{equation}
\begin{aligned}
&\hat{m}^{(k)}=\sum_{l=1}^L\frac{w^{(k)}}{G_k(v^{(l)})}\cdot F_k(v^{(l)}),\\
&\frac{1}{w^{(k)}}=\sum_{l=1}^L\frac{1}{G_k(v^{(l)})}
\end{aligned}
\end{equation}
where $\hat{m}^{(k)}$ is the aggregated mean and also the predicted $k$-th AM. However, the aggregated variance $w^{(k)}$ is not used as the uncertainty of the predicted $k$-th AM since the conditional independence assumption does not always hold in cases such as noises, silences, the computed aggregated variance will be biased by the number of voice segments in the long recording. So we calibrate the uncertainty as $\hat{w}^{(k)}=L\cdot w^{(k)}$. 

\para{Predictable AM identification.}
We have collected a number of AMs and trained estimators for predicting them. However, only a few of the AMs are actually predictable from voice, which we had anticipated while designing the task. To identify those AMs, we use hypothesis testing to them. Formally, we can write the null and alternative hypotheses for the $k$-th AM as 
\begin{itemize}
\item[] $H_0:$ the AM $m^{(k)}$ is NOT predictable from voice
\item[] $H_1:$ the AM $m^{(k)}$ is predictable from voice
\end{itemize}
In order to reject $H_0$, we only need to find a counterexample to show that voice is indeed useful in predicting AM $m^{(k)}$. An effective example is to compare the estimators with and without the voice input. If there exists a learned estimator $F_k(v)$ performing better than the chance-level estimator $C_k$ without using voice input and the results are statistically significant, we can successfully reject $H_0$ and accept $H_1$. Here the chance-level estimator for the $k$-th AM is a constant $C_k=\frac{1}{|\mathcal{D}_t|}\sum_{m^{(k)}\in\mathcal{D}_t}m^{(k)}$, which is the mean $m^{(k)}$ of the training set $\mathcal{D}_t$. So the null and alternative hypothesis can be rewritten as
\begin{itemize}
\centering
    \item[] \hspace{-1cm}$H_0:\mu(\epsilon_k/\epsilon_k^C)\leq 1$
    \item[] \hspace{-1cm}$H_1:\mu(\epsilon_k/\epsilon_k^C)\geq 1$
\end{itemize}
where $\epsilon_k$ and $\epsilon_k^C$ are the mean square errors of estimators with and without voice inputs on validation set $\mathcal{D}_{v2}$, respectively. The formulations of $\epsilon_k$ and $\epsilon_k^C$ are given as $\epsilon_k=\frac{1}{|\mathcal{D}_{v2}|}\sum_{m^{(k)}\in\mathcal{D}_{v2}}(\hat{m}^{(k)}-m^{(k)})^2$ and $\epsilon_k^C=\frac{1}{|\mathcal{D}_{v2}|}\sum_{m^{(k)}\in\mathcal{D}_{v2}}(C_k-m^{(k)})^2$.
Since the true variance of $\epsilon_k/\epsilon_k^C$ is unknown, the type of hypothesis testing is one-sided paired-sample t-test. The upper bound of the confidence interval (CI) is given by
\begin{equation}
    CI_u=\mu(\epsilon_k/\epsilon_k^C)+t_{1-\alpha,\nu}\cdot\frac{\sigma(\epsilon_k/\epsilon_k^C)}{\sqrt{N}}
\end{equation}
where $\mu(\cdot)$ and $\sigma(\cdot)$ are the functions for computing mean and standard deviation respectively. $N$ is the number of the repeated experiments and we set $N=100$ here. $\alpha$ and $\nu=N-1$ are the significance level and the degree of freedom respectively. For the purpose of this section, we adopt the significance level of 5\% and then we can read $t_{0.95,N-1}$ from t-distribution table. Now we can determine whether to reject $H_0$ and accept $H_1$,\ie, the AM $m^{(k)}$ is predictable from voice. According to the experimental results, the probability that the aforementioned decision is correct is higher than 95\%,\ie, statistically significant. In contrast, $CI_u\geq 1$ implies that we fail to reject $H_0$, for the current experimental results are not statistically significant enough. Note that failing to reject $H_0$ does not imply we accept $H_0$.

We emphasize that it is necessary to compute $\epsilon_k^C$ and $\epsilon_k$ on $\mathcal{D}_{v2}$ rather than $\mathcal{D}_t$ or $\mathcal{D}_{v1}$. This is because our estimators are trained on $\mathcal{D}_t$ and selected by the errors on $\mathcal{D}_{v1}$, we can easily get significantly lower $\epsilon_k$ and $\epsilon_k^C$ on these splits. 

\para{Optional phonatory module.}
Inspired by linear predictive coding (LPC) \cite{markel1976linear} which leverages voice producing to learn vocal tract geometry, we aim to facilitate face geometry capture by learning characteristics of voice. We enroll a phonatory module serving as an additional constraint when predicting facial AMs. In particular, we leverage a diffusion-based \cite{ho2020denoising} voice generation method to model the time-domain speech signals. As shown in \cref{fig:pipeline}, the diffusion model converts the noise distribution to a speech $\Tilde{v}$ controlled by the voice code $e$ extracted from speech $v$. During training speech $v^\prime$ which shares speaker identity with $v$ is fed to the diffusion model as ground-truth. Please note that the phonatory module only serves as an additional training constraint and is not applied during inference. Let $x_0,\cdots,x_T$ be a sequence of variables with the same dimension where $t$ is the index for diffusion time steps. Then the diffusion process transforms $x_0$ into a Gaussian noise $x_T$ through a chain of Markov transitions with a set of variance schedule $\beta_1,\cdots,\beta_T$. Specifically, each transformation is performed according to the Markov transition probability $q(x_t|x_{t-1},e)$ assumed to be independent of the style code $e$ as
\begin{equation}
    q(x_t|x_{t-1},e)=\mathcal{N}(x_t;\sqrt{1-\beta_tx_{t-1}}, \beta_tI).
\end{equation}
Unlike the diffusion process, the denoising process aims to recover the speech signal from Gaussian noise which is defined as a conditional distribution $p_\theta(x_{0:T-1}|x_T,c)$. Through the reverse transitions $p_\theta(x_{0:T-1}|x_T,c)$, the variables are gradually restored to a speech signal with style code condition. The phonatory module actually models a distribution $q(x_0|c)$. By applying the parameterization trick \cite{kingma2013auto}, we obtain the additional training constraint as
\begin{equation}
    \{\mathcal{E}^*,\theta^*\}=\argmin_{\mathcal{E},\theta}=\mathbb{E}_{x_0,\epsilon,t}\|\epsilon-\epsilon_\theta(\sqrt{\Bar{\alpha}_t}x_0+\sqrt{1-\Bar{\alpha}_t}\epsilon,t,e)\|_1
\end{equation}
where $\alpha_t=1-\beta_t$ and $\Bar{\alpha}_t=\prod^t_{t^\prime=1}\alpha_{t^\prime}$. As shown in \cref{fig:pipeline}, the $\theta$ is a Net \cite{ronneberger2015u} with cross-attention \cite{rombach2022high}. Since the phonatory model is only utilized as an auxiliary constraint during training, we omit the inference details to obtain $\Tilde{v}$ here.

\subsection{AM-Guided 3D Facial Shape Reconstruction}
To reconstruct the 3D facial shape, we first need to predict AMs of the voice recordings in $\mathcal{D}_e$ first. Subsequently, we generate the 3D facial shapes based on the predicted AMs by an optimization-based method. To do so, we first project the 3D facial shapes into a low-dimensional linear space \cite{blanz2003face}. By adjusting the coefficients in low-dimensional space, we obtain different re-projected 3D facial shapes. The learning objective is to find a set of coefficients, such that the differences between the AMs of the re-projected 3D facial shape and the predicted AMs are minimized. Specifically, we construct a big matrix $B=[b_1,b_2,\cdots]\in\mathbb{R}^{3T\times |\mathcal{D}_t|}$ where each column $b_i\in\mathbb{R}^{3T\times 1}$ is a long vector obtained by flattening a 3D facial shape $f_i\in\mathbb{R}^{T\times 3}$. $T$ is the number of vertices on 3D faces. Since $3T\gg|\mathcal{D}_t|$, we compute the project matrix $P\in\mathbb{R}^{3T\times d} (d\gg 3T)$ using eigenfaces \cite{blanz2003face} on $B$. Now any flattened 3D facial shape $b$ can be approximated by re-projecting a low-dimensional vector $\beta\in\mathbb{R}^{b\times 1}$ in the form of $P\beta$. We define the computation of AM as $Q_k(b):b\mapsto \mathbb{R}$, which maps any flattened 3D facial shape $b$ into the $k$-th AM of $b$. Since $Q_k(\cdot)$ computes a distance, a proportion, or an angle of the 3D facial shape, it is a differentiable function. The optimization objective is given below.
\begin{equation}
    \beta^*=\argmin_{\beta}\lambda\|\beta\|_2^2+\sum_{k=1}^K(Q_k(P\beta)-\hat{m}^{(k)})^2\cdot z^{(k)}
\end{equation}
where $\lambda$ is the loss weight balancing two terms. The reconstructed 3D facial shape is given by $\hat{b}=P\beta^*$. 
\section{Experiments}
In this section, we elaborate on the dataset setting, implementation details and experimental results.

\begin{figure*}[t!]
\vspace{0.2cm}
    \centering    \includegraphics[width=\linewidth]{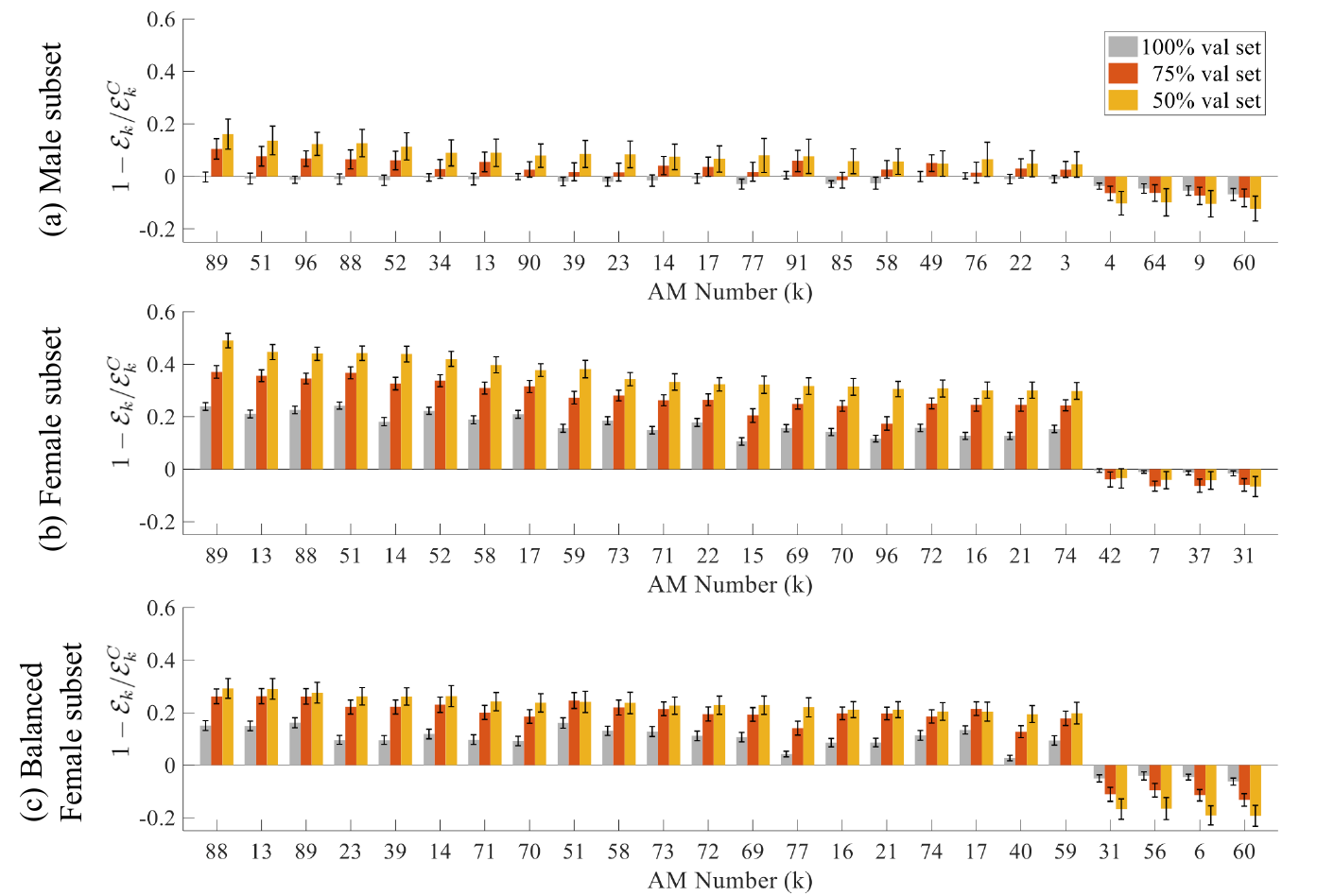}
    \caption{The normalized errors and $CI$s of 24 AMs on (a) male subset, (b) female subset, and (c) a smaller female subset. If $1-CI_u>0$, the AM is predictable else unpredictable.}
    \label{fig:AM_res}
    \vspace{0.2cm}
\end{figure*}

\subsection{Dataset}
We perform experiments on a private audiovisual dataset $\mathcal{D}$.
The dataset consists of paired voice recordings and scanned 3D facial shapes from 1,026 people, with 364 males and 662 females. The scanned 3D face is stored in the mesh format with 6790 points for each face. The voice recordings are about 2 minutes long for each speaker. We reduce the influencing factors to the voice and face by (1) asking participants to speak a set of specified sentences, (2) asking participants to speak without emotion, (3) control the age of participants (roughly 18-28 years old). In addition, to prevent the models from taking the gender shortcuts, we split the dataset $\mathcal{D}$ by gender, and experiments are individually performed on male and female subsets. For each subset, we adopt 7/1/1/1 splitting for $\mathcal{D}_t$/$\mathcal{D}_{v1}$ /$\mathcal{D}_{v2}$/$\mathcal{D}_e$. In training, the voice recordings are randomly trimmed to segments of 6 to 8 seconds, while we use the entire recordings in testing. The ground truth AMs are normalized to zero mean and unit variance. For voice features, we extract 64-dimensional log Mel-spectrograms using an analysis window of 25ms, with the hop of 10ms between frames. We perform mean and variance normalization of each Mel-frequency bin.

\subsection{Implementation Details}
We leverage a backbone $\mathcal{E}$ to learn voice code $e$ which is a simple convolutional neural network. The detailed network structure is presented in the supplementary materials. $F_k$ and $G_k$ share the backbone’s learnable parameters but have individual parameters for their heads. We use a single layer fully-connected network for each head.
For the variance head, we add an exponential activation to the last layer of $G_k$ for non-negative positive output. We follow the typical settings of stochastic gradient descent (SGD) for optimization. 
Minibatch size is 64. The momentum, learning rate, and weight decay values are 0.9, 0.1, and 0.0005, respectively. The training is completed at 5k iterations.  Since the phonatory module requires a long training procedure, we first train it with the voice code encoder $\mathcal{E}$ for 60k steps on our training set $\mathcal{D}_t$. We follow the training setting in \cite{ho2020denoising} to train the phonatory module. The other parameter setting follows \cite{ho2020denoising}. We directly normalized the voice signal as input to the network instead of first converting it to Log-Mel spectrum. To ensure statistical significance, we perform N = 100 repeated experiments to compute the $CI_u$. For the experiments at phoneme level, we leverage Wav2Vec \cite{baevski2020wav2vec} to cut the long voice recordings into phonemes.

\subsection{Predictable AM Analysis}
\label{sec:AM}
For AM prediction, the estimation models are trained on $\mathcal{D}_t$ and selected based on their performance on $\mathcal{D}_{v1}$ (hyperparameter tuning). For AM selection, the predictable AMs are selected based on the upper bound of the CI $(CI_u)$ on $\mathcal{D}_{v2}$. The performance can be evaluated by the mean error of each AM and its CI.

\cref{fig:AM_res} shows the results, including 20 AMs with highest $1-CI_u$ and 4 AMs with lowest $1-CI_u$. The gray bars are the results on the entire validation set $\mathcal{D}_{v2}$, while the red and yellow ones are the results of 75\% and 50\% voice samples with lowest uncertainty $\hat{w}$ on $\mathcal{D}_{v2}$, respectively. The self-constructed female subset has the same size as the male subset. Higher $1-CI_u$ indicates better results and the normalized error of 0 indicates the chance-level performance.  As suggested by our hypothesis testing formulation, the AMs with $1-CI_u>0$ are considered predictable from voice. In this sense, we have discovered a number of predictable female AMs (see the gray bars and their $CI_u$ in \cref{fig:AM_res} (b)). By filtering out the voice samples with high uncertainties, we achieve even higher $1-CI_u$ (see the red and yellow bars and their $CI$s). The improved performance indicates that more AMs are discovered as predictable from voice. The complete results of all AMs are given in the appendix. The results empirically demonstrate that the information of 3D facial shape is indeed encoded in the voices and can be discovered by our analysis pipeline.

\begin{table}[t]
\centering
\begin{tabular}{c|ccc}
    Phonation Module &  100\% $\hat{w}$  & 75\% $\hat{w}$ & 50\% $\hat{w}$ \\
    \hline
    \Checkmark & 0.953$\pm$ 0.009 & 0.909$\pm$0.024 & 0.842$\pm$0.030 \\
    \XSolidBrush & 0.952$\pm$0.014 & 0.927$\pm$0.030 & 0.879$\pm$0.041 \\
\end{tabular}
\caption{Effect of the phonatory module. We measure the normalized mean squared error between predicted and ground-truth AM among all AMs with different confidence thresholds.}
\label{tab:phonatory1}
\end{table}

\begin{figure}[t!]
    \centering    \includegraphics[width=\linewidth]{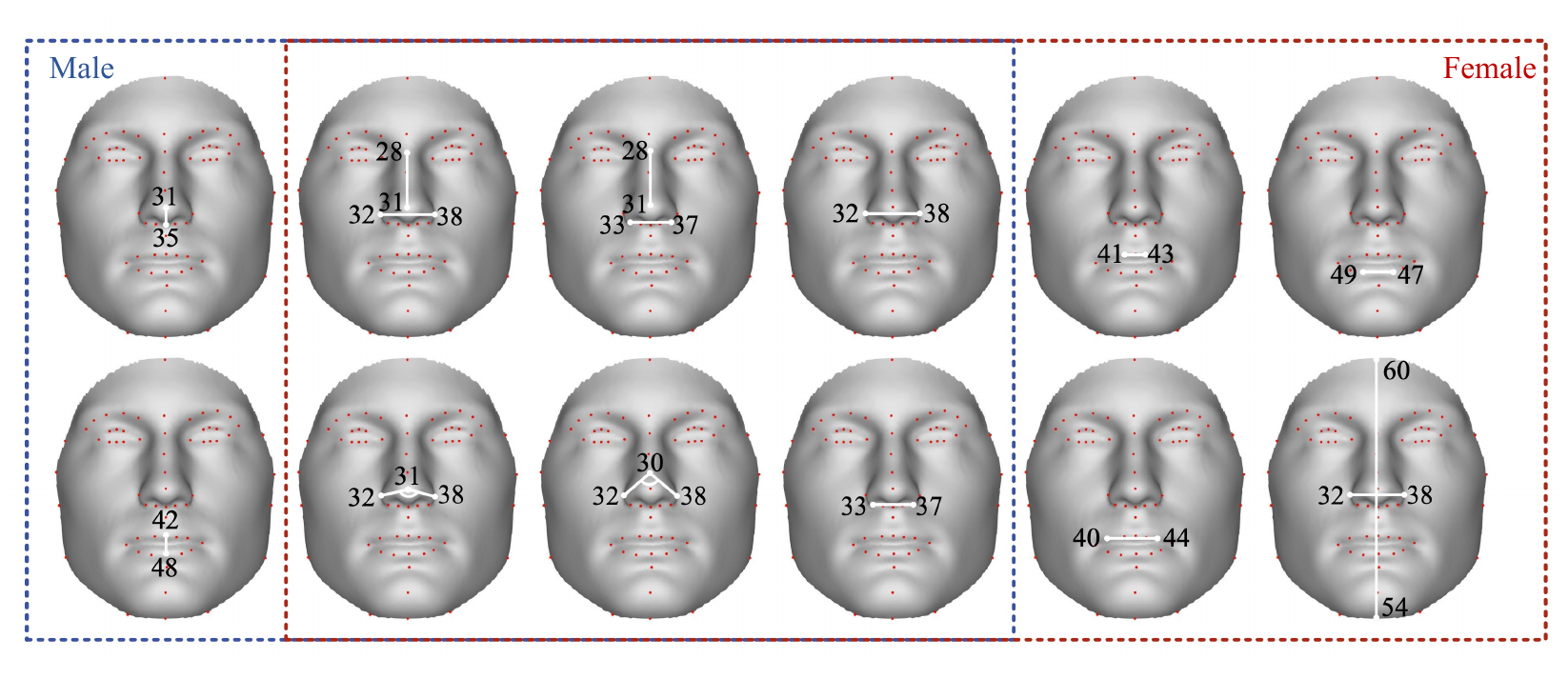}
    \caption{Visualization of the predictable AMs. Blue box: male, Red box: female.}
    \label{fig:AM_vis}
\end{figure}

To intuitively locate the predictable AMs on the 3D face, we visualize them in \cref{fig:AM_vis}. We clearly observe that most of the predictable AMs are around nose and mouth, and many of them are shared between male and female subsets. This is consistent with the fact that nose and mouth shapes affect pronunciation.

We also notice that the performance of female subset is much better than that of the male subset. To investigate whether the improvements come from the larger data scale (364 males $v.s.$ 662 females), we perform another set of repeated experiments on a self-constructed female subset, which has the same size as the male subset,\ie, 364 females. Surprisingly, the results on the new subset are still better than those on the male subset, as shown in \cref{fig:AM_res} (c). This is possible because the female subjects have higher nasalance scores on the nasal sentences \cite{vampola2020influence} among other things, which provides useful information for predicting the AMs around the nose. Here we note that our experiments have revealed that measurements around the nose are highly correlated to voice. More investigations are left for future work.

On the other hand, some AMs have not been shown to be predictable from voice. This observation suggests that voices may only associate with a few specific regions of the 3D facial shape, like the nose and mouth. For the AMs with higher errors than chance level, we do not claim they are not predictable from voice. Instead, we fail to demonstrate their predictability based on our current empirical results. The possible reasons include imperfect modeling, limited data, data noise, etc. 

\begin{table}[t]
\centering
\begin{tabular}{c|cc}
    Phonatory Module & Predictable & Unpredictable \\
    \hline
    \Checkmark & 0.628$\pm$0.021 & 0.990$\pm$0.032\\
    \XSolidBrush & 0.730$\pm$0.048 & 1.002$\pm$0.031\\

\end{tabular}
\caption{Effect of phonatory module for predictable and unpredictable AMs. We measure the normalized mean squared error between predicted and ground-truth AM among all predictable and unpredictable AMs. Interestingly, we find phonatory module only improves predictable AMs.}
\label{tab:phonatory2}
\end{table}

\subsection{Effect of Phonatory Module}
As presented in \cref{tab:phonatory1}, it is evident that utilizing the phonatory module during training enhances the accuracy of predicted AMs. Our evaluation involved computing the normalized error across all AMs with various confidence thresholds. Although the models with and without the phonatory module exhibited a marginal difference in error when evaluating all the data, the ones trained with the phonatory module showed a clear improvement in error when considering more confident samples.

Furthermore, we conducted an error evaluation for predictable and unpredictable AMs as depicted in \cref{tab:phonatory2}. We observed that utilizing the phonatory module resulted in a 0.102-point decrease in normalized error for predictable AMs, highlighting its effectiveness in improving the prediction performance. Interestingly, the phonatory module did not have any apparent effect on unpredictable AMs. Overall, the results indicate that utilizing the phonatory module during training is beneficial for predicting AMs, particularly for predictable ones.

\begin{figure*}[t!]
    \centering    \includegraphics[width=\linewidth]{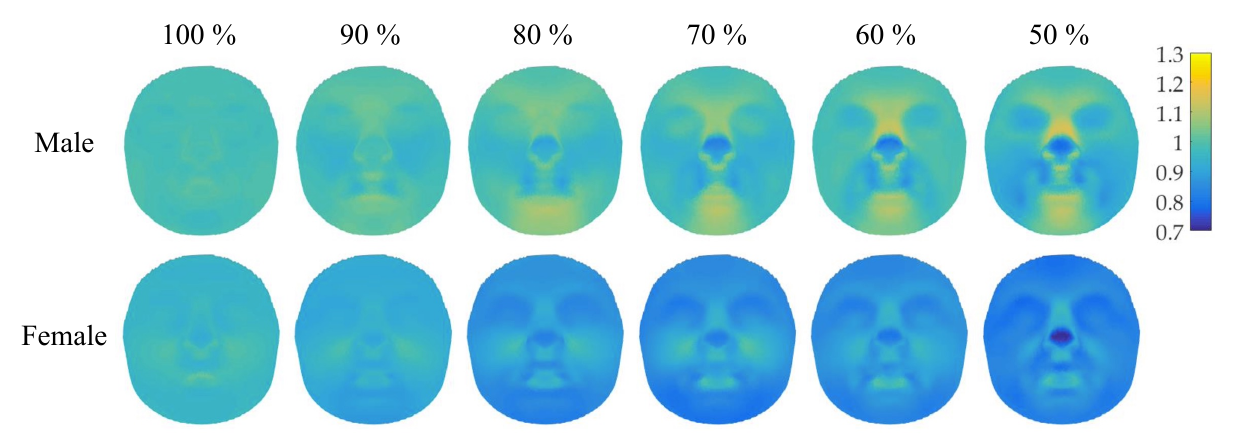}
    \caption{Error maps of the reconstructed 3D facial shapes for the male and female subsets. From left to right: the error maps corresponding to 100\% (i.e. the entire test set) to 50\% of the test set.}
    \label{fig:3d_diff}
\end{figure*}

\subsection{Phoneme-level Analysis}
\begin{figure}[h!]
    \centering\includegraphics[width=\linewidth]{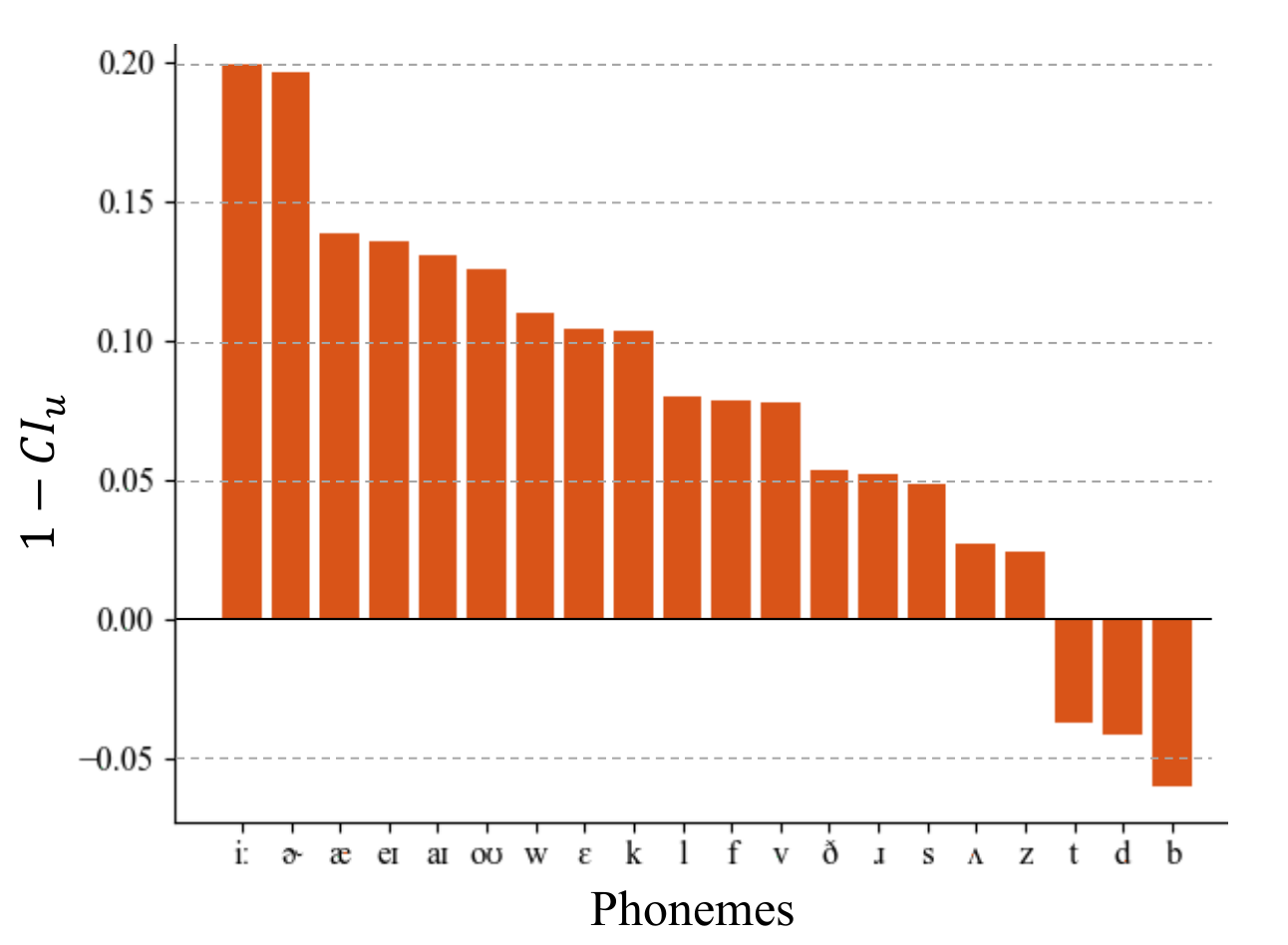}
    \caption{Phonemes with corresponding averaged $CI_u$ in decreasing order.}
    \label{fig:phoneme}
\end{figure}

We also experiment with the voice-face correlation at the phoneme level. For this experiment, we train and evaluate estimators by taking one phoneme as input each time. We computed the average $1 - CI_u$ value for each phoneme across all AMs, as shown in \cref{fig:phoneme}. Our results indicate that \texttt{/\textipa{i:}/} had the highest average $1 - CI_u$ value of 0.199, while \texttt{/\textipa{b}/} had the lowest value of -0.06. When the $1 - CI_u$ value is less than 0, it suggests that AMs are generally unpredictable from the corresponding phoneme.

We observed that the three phonemes with the lowest and negative $1 - CI_u$ values were \texttt{/\textipa{t}/}, \texttt{/\textipa{b}/}, and \texttt{/\textipa{d}/}, all of which are plosive consonants. During the pronunciation of plosive consonants, there is a complete stoppage of airflow followed by a sudden release of air through minimal mouth opening and closing. As a result, there is minimal movement of the facial muscles and structures, making it challenging for the model to predict AMs based solely on these phonemes.

In contrast, most vowels achieved good performance in the test set, with all of the top 6 phonemes belonging to vowels with $1 - CI_u > 0.10$. Compared to consonants, the production of vowels does not involve constriction of airflow in the vocal tract. Instead, the facial muscles have relatively greater movement during the pronunciation of these phonemes, such as jaw movement due to mouth opening or lip spreading. Thus, vowel phonemes may carry more information about facial features, making it easier for the model to capture the hidden correlation when predicting AMs.

\subsection{3D Facial Shape Reconstruction}
In \cref{sec:AM}, we have discovered a number of predictable AMs, from which we choose 10 AMs with the highest $1-CI_u$ for the subsequent reconstructions on male and female subsets.

To evaluate the performance, we compute the per-vertex errors between the reconstructed 3D facial shape and their ground truths. We also filter out a portion of voice samples with the highest uncertainties and evaluate the errors in the remaining data. The filter-out rate is from 0\% to 50\%, as shown from left to right in \cref{fig:3d_diff}.

Unsurprisingly, we achieve the lowest errors around the nose region for male and female subsets, consistent with the AM estimations. Moreover, the reconstruction errors decrease significantly by filtering out the voice samples with the highest uncertainties. This indicates that the learned uncertainty is effectively associated with the reconstruction quality and allows the system to decide whether to trust the model or not.

\section{Conclusion}
In conclusion, this paper presents a novel approach to exploring the voice-face correlation by focusing on the geometric aspects of the face rather than relying on semantic cues such as gender, age, and emotion. The proposed voice-anthropometric measurement (AM)-face paradigm identifies predictable facial AMs from the voice to guide 3D face reconstruction, which results in significant correlations between voice and specific parts of the face geometry, such as the nasal cavity and cranium. This approach not only eliminates the influence of unpredictable AMs but also offers a new perspective on voice-face correlation, which can be valuable for anthropometry science. The results of this study open up possibilities for future research in this area, such as developing more accurate voice-guided face synthesis techniques and a better understanding of the relationship between voice and facial geometry.

\clearpage
{\small
\bibliographystyle{ACM-Reference-Format}
\bibliography{egbib}


\begin{thebibliography}{55}


\ifx \showCODEN    \undefined \def \showCODEN     #1{\unskip}     \fi
\ifx \showDOI      \undefined \def \showDOI       #1{#1}\fi
\ifx \showISBNx    \undefined \def \showISBNx     #1{\unskip}     \fi
\ifx \showISBNxiii \undefined \def \showISBNxiii  #1{\unskip}     \fi
\ifx \showISSN     \undefined \def \showISSN      #1{\unskip}     \fi
\ifx \showLCCN     \undefined \def \showLCCN      #1{\unskip}     \fi
\ifx \shownote     \undefined \def \shownote      #1{#1}          \fi
\ifx \showarticletitle \undefined \def \showarticletitle #1{#1}   \fi
\ifx \showURL      \undefined \def \showURL       {\relax}        \fi
\providecommand\bibfield[2]{#2}
\providecommand\bibinfo[2]{#2}
\providecommand\natexlab[1]{#1}
\providecommand\showeprint[2][]{arXiv:#2}

\bibitem[Ali et~al\mbox{.}(2017)]%
        {ali2017automatic}
\bibfield{author}{\bibinfo{person}{Zulfiqar Ali}, \bibinfo{person}{Ghulam
  Muhammad}, {and} \bibinfo{person}{Mohammed~F Alhamid}.}
  \bibinfo{year}{2017}\natexlab{}.
\newblock \showarticletitle{An automatic health monitoring system for patients
  suffering from voice complications in smart cities}.
\newblock \bibinfo{journal}{\emph{IEEE Access}}  \bibinfo{volume}{5}
  (\bibinfo{year}{2017}), \bibinfo{pages}{3900--3908}.
\newblock


\bibitem[Baevski et~al\mbox{.}(2020)]%
        {baevski2020wav2vec}
\bibfield{author}{\bibinfo{person}{Alexei Baevski}, \bibinfo{person}{Yuhao
  Zhou}, \bibinfo{person}{Abdelrahman Mohamed}, {and} \bibinfo{person}{Michael
  Auli}.} \bibinfo{year}{2020}\natexlab{}.
\newblock \showarticletitle{wav2vec 2.0: A framework for self-supervised
  learning of speech representations}.
\newblock \bibinfo{journal}{\emph{Advances in neural information processing
  systems}}  \bibinfo{volume}{33} (\bibinfo{year}{2020}),
  \bibinfo{pages}{12449--12460}.
\newblock


\bibitem[Bahari et~al\mbox{.}(2012)]%
        {Bahari2012AgeEF}
\bibfield{author}{\bibinfo{person}{Mohamad~Hasan Bahari},
  \bibinfo{person}{Mitchell McLaren}, \bibinfo{person}{Hugo~Van hamme}, {and}
  \bibinfo{person}{David~A. van Leeuwen}.} \bibinfo{year}{2012}\natexlab{}.
\newblock \showarticletitle{Age Estimation from Telephone Speech using
  i-vectors}. In \bibinfo{booktitle}{\emph{Interspeech}}.
\newblock


\bibitem[Blanz and Vetter(1999)]%
        {blanz1999morphable}
\bibfield{author}{\bibinfo{person}{Volker Blanz} {and} \bibinfo{person}{Thomas
  Vetter}.} \bibinfo{year}{1999}\natexlab{}.
\newblock \showarticletitle{A morphable model for the synthesis of 3D faces}.
  In \bibinfo{booktitle}{\emph{Proceedings of the 26th annual conference on
  Computer graphics and interactive techniques}}. \bibinfo{pages}{187--194}.
\newblock


\bibitem[Blanz and Vetter(2003)]%
        {blanz2003face}
\bibfield{author}{\bibinfo{person}{Volker Blanz} {and} \bibinfo{person}{Thomas
  Vetter}.} \bibinfo{year}{2003}\natexlab{}.
\newblock \showarticletitle{Face recognition based on fitting a 3D morphable
  model}.
\newblock \bibinfo{journal}{\emph{IEEE Transactions on pattern analysis and
  machine intelligence}} \bibinfo{volume}{25}, \bibinfo{number}{9}
  (\bibinfo{year}{2003}), \bibinfo{pages}{1063--1074}.
\newblock


\bibitem[Bull et~al\mbox{.}(1983a)]%
        {bull1983voice}
\bibfield{author}{\bibinfo{person}{Ray Bull}, \bibinfo{person}{Harriet
  Rathborn}, {and} \bibinfo{person}{Brian~R Clifford}.}
  \bibinfo{year}{1983}\natexlab{a}.
\newblock \showarticletitle{The voice-recognition accuracy of blind listeners}.
\newblock \bibinfo{journal}{\emph{Perception}} \bibinfo{volume}{12},
  \bibinfo{number}{2} (\bibinfo{year}{1983}), \bibinfo{pages}{223--226}.
\newblock


\bibitem[Bull et~al\mbox{.}(1983b)]%
        {Bull1983TheVA}
\bibfield{author}{\bibinfo{person}{R.~H.~C. Bull}, \bibinfo{person}{Harriet
  Rathborn}, {and} \bibinfo{person}{Brian~R. Clifford}.}
  \bibinfo{year}{1983}\natexlab{b}.
\newblock \showarticletitle{The Voice-Recognition Accuracy of Blind Listeners}.
\newblock \bibinfo{journal}{\emph{Perception}}  \bibinfo{volume}{12}
  (\bibinfo{year}{1983}), \bibinfo{pages}{223 -- 226}.
\newblock


\bibitem[Chen et~al\mbox{.}(2018)]%
        {chen2018lip}
\bibfield{author}{\bibinfo{person}{Lele Chen}, \bibinfo{person}{Zhiheng Li},
  \bibinfo{person}{Ross~K Maddox}, \bibinfo{person}{Zhiyao Duan}, {and}
  \bibinfo{person}{Chenliang Xu}.} \bibinfo{year}{2018}\natexlab{}.
\newblock \showarticletitle{Lip movements generation at a glance}. In
  \bibinfo{booktitle}{\emph{ECCV}}. \bibinfo{pages}{520--535}.
\newblock


\bibitem[Cudeiro et~al\mbox{.}(2019)]%
        {cudeiro2019capture}
\bibfield{author}{\bibinfo{person}{Daniel Cudeiro}, \bibinfo{person}{Timo
  Bolkart}, \bibinfo{person}{Cassidy Laidlaw}, \bibinfo{person}{Anurag Ranjan},
  {and} \bibinfo{person}{Michael~J Black}.} \bibinfo{year}{2019}\natexlab{}.
\newblock \showarticletitle{Capture, learning, and synthesis of 3D speaking
  styles}. In \bibinfo{booktitle}{\emph{CVPR}}. \bibinfo{pages}{10101--10111}.
\newblock


\bibitem[Farkas et~al\mbox{.}(2004)]%
        {farkas2004anthropometric}
\bibfield{author}{\bibinfo{person}{Leslie~G Farkas}, \bibinfo{person}{Otto~G
  Eiben}, \bibinfo{person}{Stefan Sivkov}, \bibinfo{person}{Bryan Tompson},
  \bibinfo{person}{Marko~J Katic}, {and} \bibinfo{person}{Christopher~R
  Forrest}.} \bibinfo{year}{2004}\natexlab{}.
\newblock \showarticletitle{Anthropometric measurements of the facial framework
  in adulthood: age-related changes in eight age categories in 600 healthy
  white North Americans of European ancestry from 16 to 90 years of age}.
\newblock \bibinfo{journal}{\emph{Journal of Craniofacial Surgery}}
  \bibinfo{volume}{15}, \bibinfo{number}{2} (\bibinfo{year}{2004}),
  \bibinfo{pages}{288--298}.
\newblock


\bibitem[Ghafourzadeh et~al\mbox{.}(2019)]%
        {ghafourzadeh2019part}
\bibfield{author}{\bibinfo{person}{Donya Ghafourzadeh}, \bibinfo{person}{Cyrus
  Rahgoshay}, \bibinfo{person}{Sahel Fallahdoust}, \bibinfo{person}{Adeline
  Aubame}, \bibinfo{person}{Andre Beauchamp}, \bibinfo{person}{Tiberiu Popa},
  {and} \bibinfo{person}{Eric Paquette}.} \bibinfo{year}{2019}\natexlab{}.
\newblock \showarticletitle{Part-based 3D face morphable model with
  anthropometric local control}.
\newblock  (\bibinfo{year}{2019}).
\newblock


\bibitem[Ghosh and Narayanan(2011)]%
        {ghosh2011automatic}
\bibfield{author}{\bibinfo{person}{Prasanta~Kumar Ghosh} {and}
  \bibinfo{person}{Shrikanth Narayanan}.} \bibinfo{year}{2011}\natexlab{}.
\newblock \showarticletitle{Automatic speech recognition using articulatory
  features from subject-independent acoustic-to-articulatory inversion}.
\newblock \bibinfo{journal}{\emph{The Journal of the Acoustical Society of
  America}} \bibinfo{volume}{130}, \bibinfo{number}{4} (\bibinfo{year}{2011}),
  \bibinfo{pages}{EL251--EL257}.
\newblock


\bibitem[Gomez and Danuser(2007)]%
        {gomez2007relationships}
\bibfield{author}{\bibinfo{person}{Patrick Gomez} {and}
  \bibinfo{person}{Brigitta Danuser}.} \bibinfo{year}{2007}\natexlab{}.
\newblock \showarticletitle{Relationships between musical structure and
  psychophysiological measures of emotion.}
\newblock \bibinfo{journal}{\emph{Emotion}} \bibinfo{volume}{7},
  \bibinfo{number}{2} (\bibinfo{year}{2007}), \bibinfo{pages}{377}.
\newblock


\bibitem[Goodfellow et~al\mbox{.}(2020)]%
        {goodfellow2020generative}
\bibfield{author}{\bibinfo{person}{Ian Goodfellow}, \bibinfo{person}{Jean
  Pouget-Abadie}, \bibinfo{person}{Mehdi Mirza}, \bibinfo{person}{Bing Xu},
  \bibinfo{person}{David Warde-Farley}, \bibinfo{person}{Sherjil Ozair},
  \bibinfo{person}{Aaron Courville}, {and} \bibinfo{person}{Yoshua Bengio}.}
  \bibinfo{year}{2020}\natexlab{}.
\newblock \showarticletitle{Generative adversarial networks}.
\newblock \bibinfo{journal}{\emph{Commun. ACM}} \bibinfo{volume}{63},
  \bibinfo{number}{11} (\bibinfo{year}{2020}), \bibinfo{pages}{139--144}.
\newblock


\bibitem[Grzybowska and Kacprzak(2016)]%
        {grzybowska2016speaker}
\bibfield{author}{\bibinfo{person}{Joanna Grzybowska} {and}
  \bibinfo{person}{Stanislaw Kacprzak}.} \bibinfo{year}{2016}\natexlab{}.
\newblock \showarticletitle{Speaker Age Classification and Regression Using
  i-Vectors.}. In \bibinfo{booktitle}{\emph{INTERSPEECH}}.
  \bibinfo{pages}{1402--1406}.
\newblock


\bibitem[Guo et~al\mbox{.}(2021)]%
        {guo2021adnerf}
\bibfield{author}{\bibinfo{person}{Yudong Guo}, \bibinfo{person}{Keyu Chen},
  \bibinfo{person}{Sen Liang}, \bibinfo{person}{Yongjin Liu},
  \bibinfo{person}{Hujun Bao}, {and} \bibinfo{person}{Juyong Zhang}.}
  \bibinfo{year}{2021}\natexlab{}.
\newblock \showarticletitle{AD-NeRF: Audio Driven Neural Radiance Fields for
  Talking Head Synthesis}. In \bibinfo{booktitle}{\emph{ICCV}}.
\newblock


\bibitem[Han et~al\mbox{.}(2021)]%
        {han2021exploring}
\bibfield{author}{\bibinfo{person}{Jing Han}, \bibinfo{person}{Chlo{\"e}
  Brown}, \bibinfo{person}{Jagmohan Chauhan}, \bibinfo{person}{Andreas
  Grammenos}, \bibinfo{person}{Apinan Hasthanasombat},
  \bibinfo{person}{Dimitris Spathis}, \bibinfo{person}{Tong Xia},
  \bibinfo{person}{Pietro Cicuta}, {and} \bibinfo{person}{Cecilia Mascolo}.}
  \bibinfo{year}{2021}\natexlab{}.
\newblock \showarticletitle{Exploring Automatic COVID-19 Diagnosis via voice
  and symptoms from Crowdsourced Data}. In \bibinfo{booktitle}{\emph{ICASSP}}.
  IEEE.
\newblock


\bibitem[Ho et~al\mbox{.}(2020)]%
        {ho2020denoising}
\bibfield{author}{\bibinfo{person}{Jonathan Ho}, \bibinfo{person}{Ajay Jain},
  {and} \bibinfo{person}{Pieter Abbeel}.} \bibinfo{year}{2020}\natexlab{}.
\newblock \showarticletitle{Denoising diffusion probabilistic models}.
\newblock \bibinfo{journal}{\emph{Advances in Neural Information Processing
  Systems}}  \bibinfo{volume}{33} (\bibinfo{year}{2020}),
  \bibinfo{pages}{6840--6851}.
\newblock


\bibitem[Jamaludin et~al\mbox{.}(2019)]%
        {jamaludin2019you}
\bibfield{author}{\bibinfo{person}{Amir Jamaludin}, \bibinfo{person}{Joon~Son
  Chung}, {and} \bibinfo{person}{Andrew Zisserman}.}
  \bibinfo{year}{2019}\natexlab{}.
\newblock \showarticletitle{You said that?: Synthesising talking faces from
  audio}.
\newblock \bibinfo{journal}{\emph{International Journal of Computer Vision
  (IJCV)}} \bibinfo{volume}{127}, \bibinfo{number}{11} (\bibinfo{year}{2019}),
  \bibinfo{pages}{1767--1779}.
\newblock


\bibitem[Kendall and Gal(2017)]%
        {kendall2017uncertainties}
\bibfield{author}{\bibinfo{person}{Alex Kendall} {and} \bibinfo{person}{Yarin
  Gal}.} \bibinfo{year}{2017}\natexlab{}.
\newblock \showarticletitle{What uncertainties do we need in bayesian deep
  learning for computer vision?}
\newblock \bibinfo{journal}{\emph{Advances in neural information processing
  systems}}  \bibinfo{volume}{30} (\bibinfo{year}{2017}).
\newblock


\bibitem[Kingma and Welling(2013)]%
        {kingma2013auto}
\bibfield{author}{\bibinfo{person}{Diederik~P Kingma} {and}
  \bibinfo{person}{Max Welling}.} \bibinfo{year}{2013}\natexlab{}.
\newblock \showarticletitle{Auto-encoding variational bayes}.
\newblock \bibinfo{journal}{\emph{arXiv preprint arXiv:1312.6114}}
  (\bibinfo{year}{2013}).
\newblock


\bibitem[Li et~al\mbox{.}(2019a)]%
        {li2019improving}
\bibfield{author}{\bibinfo{person}{Sheng Li}, \bibinfo{person}{Dabre Raj},
  \bibinfo{person}{Xugang Lu}, \bibinfo{person}{Peng Shen},
  \bibinfo{person}{Tatsuya Kawahara}, {and} \bibinfo{person}{Hisashi Kawai}.}
  \bibinfo{year}{2019}\natexlab{a}.
\newblock \showarticletitle{Improving Transformer-Based Speech Recognition
  Systems with Compressed Structure and Speech Attributes Augmentation.}. In
  \bibinfo{booktitle}{\emph{INTERSPEECH}}. \bibinfo{pages}{4400--4404}.
\newblock


\bibitem[Li et~al\mbox{.}(2019b)]%
        {Li2019ImprovingTS}
\bibfield{author}{\bibinfo{person}{Sheng Li}, \bibinfo{person}{Dabre Raj},
  \bibinfo{person}{Xugang Lu}, \bibinfo{person}{Peng Shen},
  \bibinfo{person}{Tatsuya Kawahara}, {and} \bibinfo{person}{Hisashi Kawai}.}
  \bibinfo{year}{2019}\natexlab{b}.
\newblock \showarticletitle{Improving Transformer-Based Speech Recognition
  Systems with Compressed Structure and Speech Attributes Augmentation}. In
  \bibinfo{booktitle}{\emph{Interspeech}}.
\newblock


\bibitem[Maguinness et~al\mbox{.}(2018)]%
        {maguinness2018understanding}
\bibfield{author}{\bibinfo{person}{Corrina Maguinness},
  \bibinfo{person}{Claudia Roswandowitz}, {and} \bibinfo{person}{Katharina von
  Kriegstein}.} \bibinfo{year}{2018}\natexlab{}.
\newblock \showarticletitle{Understanding the mechanisms of familiar
  voice-identity recognition in the human brain}.
\newblock \bibinfo{journal}{\emph{Neuropsychologia}}  \bibinfo{volume}{116}
  (\bibinfo{year}{2018}), \bibinfo{pages}{179--193}.
\newblock


\bibitem[Markel et~al\mbox{.}(1976)]%
        {markel1976linear}
\bibfield{author}{\bibinfo{person}{John~D Markel}, \bibinfo{person}{Augustine~H
  Gray}, {and} \bibinfo{person}{Augustine~H Gray}.}
  \bibinfo{year}{1976}\natexlab{}.
\newblock \showarticletitle{Linear prediction of speech: Communication and
  cybernetics}.
\newblock  (\bibinfo{year}{1976}).
\newblock


\bibitem[Mirza and Osindero(2014)]%
        {mirza2014conditional}
\bibfield{author}{\bibinfo{person}{Mehdi Mirza} {and} \bibinfo{person}{Simon
  Osindero}.} \bibinfo{year}{2014}\natexlab{}.
\newblock \showarticletitle{Conditional generative adversarial nets}.
\newblock \bibinfo{journal}{\emph{arXiv preprint arXiv:1411.1784}}
  (\bibinfo{year}{2014}).
\newblock


\bibitem[Nawaz et~al\mbox{.}(2021)]%
        {Nawaz_2021_CVPR}
\bibfield{author}{\bibinfo{person}{Shah Nawaz}, \bibinfo{person}{Muhammad~Saad
  Saeed}, \bibinfo{person}{Pietro Morerio}, \bibinfo{person}{Arif Mahmood},
  \bibinfo{person}{Ignazio Gallo}, \bibinfo{person}{Muhammad~Haroon Yousaf},
  {and} \bibinfo{person}{Alessio Del~Bue}.} \bibinfo{year}{2021}\natexlab{}.
\newblock \showarticletitle{Cross-Modal Speaker Verification and Recognition: A
  Multilingual Perspective}. In \bibinfo{booktitle}{\emph{CVPRW}}.
\newblock


\bibitem[Ning et~al\mbox{.}(2021)]%
        {ning2021disentangled}
\bibfield{author}{\bibinfo{person}{Hailong Ning}, \bibinfo{person}{Xiangtao
  Zheng}, \bibinfo{person}{Xiaoqiang Lu}, {and} \bibinfo{person}{Yuan Yuan}.}
  \bibinfo{year}{2021}\natexlab{}.
\newblock \showarticletitle{Disentangled Representation Learning for
  Cross-modal Biometric Matching}.
\newblock \bibinfo{journal}{\emph{TMM}} (\bibinfo{year}{2021}).
\newblock


\bibitem[Oh et~al\mbox{.}(2019)]%
        {oh2019speech2face}
\bibfield{author}{\bibinfo{person}{Tae-Hyun Oh}, \bibinfo{person}{Tali Dekel},
  \bibinfo{person}{Changil Kim}, \bibinfo{person}{Inbar Mosseri},
  \bibinfo{person}{William~T Freeman}, \bibinfo{person}{Michael Rubinstein},
  {and} \bibinfo{person}{Wojciech Matusik}.} \bibinfo{year}{2019}\natexlab{}.
\newblock \showarticletitle{Speech2face: Learning the face behind a voice}. In
  \bibinfo{booktitle}{\emph{Proceedings of the IEEE/CVF conference on computer
  vision and pattern recognition}}. \bibinfo{pages}{7539--7548}.
\newblock


\bibitem[Ptacek and Sander(1966a)]%
        {ptacek1966age}
\bibfield{author}{\bibinfo{person}{Paul~H Ptacek} {and} \bibinfo{person}{Eric~K
  Sander}.} \bibinfo{year}{1966}\natexlab{a}.
\newblock \showarticletitle{Age recognition from voice}.
\newblock \bibinfo{journal}{\emph{Journal of speech and hearing Research}}
  \bibinfo{volume}{9}, \bibinfo{number}{2} (\bibinfo{year}{1966}),
  \bibinfo{pages}{273--277}.
\newblock


\bibitem[Ptacek and Sander(1966b)]%
        {Ptacek1966AgeRF}
\bibfield{author}{\bibinfo{person}{Paul~H. Ptacek} {and}
  \bibinfo{person}{Eric~K. Sander}.} \bibinfo{year}{1966}\natexlab{b}.
\newblock \showarticletitle{Age recognition from voice.}
\newblock \bibinfo{journal}{\emph{Journal of speech and hearing research}}
  \bibinfo{volume}{9 2} (\bibinfo{year}{1966}), \bibinfo{pages}{273--7}.
\newblock


\bibitem[Ramanathan and Chellappa(2006)]%
        {ramanathan2006modeling}
\bibfield{author}{\bibinfo{person}{Narayanan Ramanathan} {and}
  \bibinfo{person}{Rama Chellappa}.} \bibinfo{year}{2006}\natexlab{}.
\newblock \showarticletitle{Modeling age progression in young faces}. In
  \bibinfo{booktitle}{\emph{2006 IEEE Computer Society Conference on Computer
  Vision and Pattern Recognition (CVPR'06)}}, Vol.~\bibinfo{volume}{1}. IEEE,
  \bibinfo{pages}{387--394}.
\newblock


\bibitem[Ravanelli and Bengio(2018a)]%
        {ravanelli2018speaker}
\bibfield{author}{\bibinfo{person}{Mirco Ravanelli} {and}
  \bibinfo{person}{Yoshua Bengio}.} \bibinfo{year}{2018}\natexlab{a}.
\newblock \showarticletitle{Speaker recognition from raw waveform with
  sincnet}. In \bibinfo{booktitle}{\emph{2018 IEEE Spoken Language Technology
  Workshop (SLT)}}. IEEE, \bibinfo{pages}{1021--1028}.
\newblock


\bibitem[Ravanelli and Bengio(2018b)]%
        {Ravanelli2018SpeakerRF}
\bibfield{author}{\bibinfo{person}{Mirco Ravanelli} {and}
  \bibinfo{person}{Yoshua Bengio}.} \bibinfo{year}{2018}\natexlab{b}.
\newblock \showarticletitle{Speaker Recognition from Raw Waveform with
  SincNet}.
\newblock \bibinfo{journal}{\emph{2018 IEEE Spoken Language Technology Workshop
  (SLT)}} (\bibinfo{year}{2018}), \bibinfo{pages}{1021--1028}.
\newblock


\bibitem[Rombach et~al\mbox{.}(2022)]%
        {rombach2022high}
\bibfield{author}{\bibinfo{person}{Robin Rombach}, \bibinfo{person}{Andreas
  Blattmann}, \bibinfo{person}{Dominik Lorenz}, \bibinfo{person}{Patrick
  Esser}, {and} \bibinfo{person}{Bj{\"o}rn Ommer}.}
  \bibinfo{year}{2022}\natexlab{}.
\newblock \showarticletitle{High-resolution image synthesis with latent
  diffusion models}. In \bibinfo{booktitle}{\emph{Proceedings of the IEEE/CVF
  Conference on Computer Vision and Pattern Recognition}}.
  \bibinfo{pages}{10684--10695}.
\newblock


\bibitem[Ronneberger et~al\mbox{.}(2015)]%
        {ronneberger2015u}
\bibfield{author}{\bibinfo{person}{Olaf Ronneberger}, \bibinfo{person}{Philipp
  Fischer}, {and} \bibinfo{person}{Thomas Brox}.}
  \bibinfo{year}{2015}\natexlab{}.
\newblock \showarticletitle{U-net: Convolutional networks for biomedical image
  segmentation}. In \bibinfo{booktitle}{\emph{Medical Image Computing and
  Computer-Assisted Intervention--MICCAI 2015: 18th International Conference,
  Munich, Germany, October 5-9, 2015, Proceedings, Part III 18}}. Springer,
  \bibinfo{pages}{234--241}.
\newblock


\bibitem[Sar{\i} et~al\mbox{.}(2021)]%
        {sari2021multi}
\bibfield{author}{\bibinfo{person}{Leda Sar{\i}}, \bibinfo{person}{Kritika
  Singh}, \bibinfo{person}{Jiatong Zhou}, \bibinfo{person}{Lorenzo Torresani},
  \bibinfo{person}{Nayan Singhal}, {and} \bibinfo{person}{Yatharth Saraf}.}
  \bibinfo{year}{2021}\natexlab{}.
\newblock \showarticletitle{A Multi-View Approach to Audio-Visual Speaker
  Verification}. In \bibinfo{booktitle}{\emph{ICASSP}}.
\newblock


\bibitem[Shan et~al\mbox{.}(2021)]%
        {shan2021anthropometric}
\bibfield{author}{\bibinfo{person}{Zhiyi Shan}, \bibinfo{person}{Richard
  Tai-Chiu Hsung}, \bibinfo{person}{Congyi Zhang}, \bibinfo{person}{Juanjuan
  Ji}, \bibinfo{person}{Wing~Shan Choi}, \bibinfo{person}{Wenping Wang},
  \bibinfo{person}{Yanqi Yang}, \bibinfo{person}{Min Gu}, {and}
  \bibinfo{person}{Balvinder~S Khambay}.} \bibinfo{year}{2021}\natexlab{}.
\newblock \showarticletitle{Anthropometric﻿ accuracy of three-dimensional
  average faces compared to conventional facial measurements}.
\newblock \bibinfo{journal}{\emph{Scientific Reports}} \bibinfo{volume}{11},
  \bibinfo{number}{1} (\bibinfo{year}{2021}), \bibinfo{pages}{1--12}.
\newblock


\bibitem[Singh et~al\mbox{.}(2016a)]%
        {singh2016relationship}
\bibfield{author}{\bibinfo{person}{Rita Singh}, \bibinfo{person}{Joseph
  Keshet}, \bibinfo{person}{Deniz Gencaga}, {and} \bibinfo{person}{Bhiksha
  Raj}.} \bibinfo{year}{2016}\natexlab{a}.
\newblock \showarticletitle{The relationship of voice onset time and voice
  offset time to physical age}. In \bibinfo{booktitle}{\emph{ICASSP}}. IEEE,
  \bibinfo{pages}{5390--5394}.
\newblock


\bibitem[Singh et~al\mbox{.}(2016b)]%
        {singh2016forensic}
\bibfield{author}{\bibinfo{person}{Rita Singh}, \bibinfo{person}{Bhiksha Raj},
  {and} \bibinfo{person}{Deniz Gencaga}.} \bibinfo{year}{2016}\natexlab{b}.
\newblock \showarticletitle{Forensic anthropometry from voice: an
  articulatory-phonetic approach}. In \bibinfo{booktitle}{\emph{2016 39th
  International Convention on Information and Communication Technology,
  Electronics and Microelectronics (MIPRO)}}. IEEE,
  \bibinfo{pages}{1375--1380}.
\newblock


\bibitem[Tao et~al\mbox{.}(2020)]%
        {tao2020audio}
\bibfield{author}{\bibinfo{person}{Ruijie Tao}, \bibinfo{person}{Rohan~Kumar
  Das}, {and} \bibinfo{person}{Haizhou Li}.} \bibinfo{year}{2020}\natexlab{}.
\newblock \showarticletitle{Audio-visual speaker recognition with a cross-modal
  discriminative network}. In \bibinfo{booktitle}{\emph{INTERSPEECH}}.
\newblock


\bibitem[Vampola et~al\mbox{.}(2020)]%
        {vampola2020influence}
\bibfield{author}{\bibinfo{person}{Tom{\'a}{\v{s}} Vampola},
  \bibinfo{person}{Jarom{\'\i}r Hor{\'a}{\v{c}}ek},
  \bibinfo{person}{Vojt{\v{e}}ch Radolf}, \bibinfo{person}{Jan~G {\v{S}}vec},
  {and} \bibinfo{person}{Anne-Maria Laukkanen}.}
  \bibinfo{year}{2020}\natexlab{}.
\newblock \showarticletitle{Influence of nasal cavities on voice quality:
  Computer simulations and experiments}.
\newblock \bibinfo{journal}{\emph{The Journal of the Acoustical Society of
  America}} \bibinfo{volume}{148}, \bibinfo{number}{5} (\bibinfo{year}{2020}),
  \bibinfo{pages}{3218--3231}.
\newblock


\bibitem[Wang and Tashev(2017)]%
        {wang2017learning}
\bibfield{author}{\bibinfo{person}{Zhong-Qiu Wang} {and} \bibinfo{person}{Ivan
  Tashev}.} \bibinfo{year}{2017}\natexlab{}.
\newblock \showarticletitle{Learning utterance-level representations for speech
  emotion and age/gender recognition using deep neural networks}. In
  \bibinfo{booktitle}{\emph{ICASSP}}. IEEE, \bibinfo{pages}{5150--5154}.
\newblock


\bibitem[Wen et~al\mbox{.}(2021)]%
        {Wen_2021_CVPR}
\bibfield{author}{\bibinfo{person}{Peisong Wen}, \bibinfo{person}{Qianqian Xu},
  \bibinfo{person}{Yangbangyan Jiang}, \bibinfo{person}{Zhiyong Yang},
  \bibinfo{person}{Yuan He}, {and} \bibinfo{person}{Qingming Huang}.}
  \bibinfo{year}{2021}\natexlab{}.
\newblock \showarticletitle{Seeking the Shape of Sound: An Adaptive Framework
  for Learning Voice-Face Association}. In \bibinfo{booktitle}{\emph{CVPR}}.
  \bibinfo{pages}{16347--16356}.
\newblock


\bibitem[Wen et~al\mbox{.}(2019)]%
        {NEURIPS2019_eb9fc349}
\bibfield{author}{\bibinfo{person}{Yandong Wen}, \bibinfo{person}{Bhiksha Raj},
  {and} \bibinfo{person}{Rita Singh}.} \bibinfo{year}{2019}\natexlab{}.
\newblock \showarticletitle{Face Reconstruction from Voice using Generative
  Adversarial Networks}. In \bibinfo{booktitle}{\emph{NeurIPS}},
  Vol.~\bibinfo{volume}{32}.
\newblock


\bibitem[Wiles et~al\mbox{.}(2018)]%
        {wiles2018x2face}
\bibfield{author}{\bibinfo{person}{Olivia Wiles}, \bibinfo{person}{A Koepke},
  {and} \bibinfo{person}{Andrew Zisserman}.} \bibinfo{year}{2018}\natexlab{}.
\newblock \showarticletitle{X2face: A network for controlling face generation
  using images, audio, and pose codes}. In \bibinfo{booktitle}{\emph{ECCV}}.
  \bibinfo{pages}{670--686}.
\newblock


\bibitem[Wu et~al\mbox{.}(2022)]%
        {wu2022cross}
\bibfield{author}{\bibinfo{person}{Cho-Ying Wu}, \bibinfo{person}{Chin-Cheng
  Hsu}, {and} \bibinfo{person}{Ulrich Neumann}.}
  \bibinfo{year}{2022}\natexlab{}.
\newblock \showarticletitle{Cross-Modal Perceptionist: Can Face Geometry be
  Gleaned from Voices?}. In \bibinfo{booktitle}{\emph{Proceedings of the
  IEEE/CVF Conference on Computer Vision and Pattern Recognition}}.
  \bibinfo{pages}{10452--10461}.
\newblock


\bibitem[Wyganowska-{S}wi{k{a}}tkowska et~al\mbox{.}(2017)]%
        {wyganowska2017vocal}
\bibfield{author}{\bibinfo{person}{Marzena Wyganowska-{S}wi{k{a}}tkowska},
  \bibinfo{person}{Iwona Kowalkowska}, \bibinfo{person}{Gra{z}yna
  Flici{n}ska-Pamfil}, \bibinfo{person}{Miko{l}aj D{{a}}browski},
  \bibinfo{person}{Przemys{l}aw Kopczy{n}ski}, {and} \bibinfo{person}{Bo{z}ena
  Wiskirska-Wo{z}nica}.} \bibinfo{year}{2017}\natexlab{}.
\newblock \showarticletitle{Vocal training in an anthropometrical aspect}.
\newblock \bibinfo{journal}{\emph{Logopedics Phoniatrics Vocology}}
  \bibinfo{volume}{42}, \bibinfo{number}{4} (\bibinfo{year}{2017}),
  \bibinfo{pages}{178--186}.
\newblock


\bibitem[Wyganowska-{S}wi{k{a}}tkowska et~al\mbox{.}(2013)]%
        {wyganowska2013anthropometric}
\bibfield{author}{\bibinfo{person}{Marzena Wyganowska-{S}wi{k{a}}tkowska},
  \bibinfo{person}{Iwona Kowalkowska}, \bibinfo{person}{Katarzyna Mehr}, {and}
  \bibinfo{person}{Miko{l}aj D{k{a}}browski}.} \bibinfo{year}{2013}\natexlab{}.
\newblock \showarticletitle{An anthropometric analysis of the head and face in
  vocal students}.
\newblock \bibinfo{journal}{\emph{Folia Phoniatrica et Logopaedica}}
  \bibinfo{volume}{65}, \bibinfo{number}{3} (\bibinfo{year}{2013}),
  \bibinfo{pages}{136--142}.
\newblock


\bibitem[Yang et~al\mbox{.}(2023)]%
        {yang2023paaploss}
\bibfield{author}{\bibinfo{person}{Muqiao Yang}, \bibinfo{person}{Joseph
  Konan}, \bibinfo{person}{David Bick}, \bibinfo{person}{Yunyang Zeng},
  \bibinfo{person}{Shuo Han}, \bibinfo{person}{Anurag Kumar},
  \bibinfo{person}{Shinji Watanabe}, {and} \bibinfo{person}{Bhiksha Raj}.}
  \bibinfo{year}{2023}\natexlab{}.
\newblock \showarticletitle{PAAPLoss: A Phonetic-Aligned Acoustic Parameter
  Loss for Speech Enhancement}.
\newblock \bibinfo{journal}{\emph{Proc. of ICASSP}} (\bibinfo{year}{2023}).
\newblock


\bibitem[Zhang et~al\mbox{.}(2019a)]%
        {zhang2019attention}
\bibfield{author}{\bibinfo{person}{Zixing Zhang}, \bibinfo{person}{Bingwen Wu},
  {and} \bibinfo{person}{Bj{\"o}rn Schuller}.}
  \bibinfo{year}{2019}\natexlab{a}.
\newblock \showarticletitle{Attention-augmented end-to-end multi-task learning
  for emotion prediction from speech}. In \bibinfo{booktitle}{\emph{ICASSP}}.
  IEEE, \bibinfo{pages}{6705--6709}.
\newblock


\bibitem[Zhang et~al\mbox{.}(2019b)]%
        {Zhang2019AttentionaugmentedEM}
\bibfield{author}{\bibinfo{person}{Zixing Zhang}, \bibinfo{person}{Bingwen Wu},
  {and} \bibinfo{person}{Bj{\"o}rn Schuller}.}
  \bibinfo{year}{2019}\natexlab{b}.
\newblock \showarticletitle{Attention-augmented End-to-end Multi-task Learning
  for Emotion Prediction from Speech}.
\newblock \bibinfo{journal}{\emph{ICASSP 2019 - 2019 IEEE International
  Conference on Acoustics, Speech and Signal Processing (ICASSP)}}
  (\bibinfo{year}{2019}), \bibinfo{pages}{6705--6709}.
\newblock


\bibitem[Zheng et~al\mbox{.}(2021)]%
        {zheng2021adversarial}
\bibfield{author}{\bibinfo{person}{Aihua Zheng}, \bibinfo{person}{Menglan Hu},
  \bibinfo{person}{Bo Jiang}, \bibinfo{person}{Yan Huang}, \bibinfo{person}{Yan
  Yan}, {and} \bibinfo{person}{Bin Luo}.} \bibinfo{year}{2021}\natexlab{}.
\newblock \showarticletitle{Adversarial-metric learning for audio-visual
  cross-modal matching}.
\newblock \bibinfo{journal}{\emph{TMM}} (\bibinfo{year}{2021}).
\newblock


\bibitem[Zhou et~al\mbox{.}(2019)]%
        {zhou2019talking}
\bibfield{author}{\bibinfo{person}{Hang Zhou}, \bibinfo{person}{Yu Liu},
  \bibinfo{person}{Ziwei Liu}, \bibinfo{person}{Ping Luo}, {and}
  \bibinfo{person}{Xiaogang Wang}.} \bibinfo{year}{2019}\natexlab{}.
\newblock \showarticletitle{Talking face generation by adversarially
  disentangled audio-visual representation}. In
  \bibinfo{booktitle}{\emph{AAAI}}, Vol.~\bibinfo{volume}{33}.
  \bibinfo{pages}{9299--9306}.
\newblock


\bibitem[Zhuang et~al\mbox{.}(2010)]%
        {zhuang2010facial}
\bibfield{author}{\bibinfo{person}{Ziqing Zhuang}, \bibinfo{person}{Douglas
  Landsittel}, \bibinfo{person}{Stacey Benson}, \bibinfo{person}{Raymond
  Roberge}, {and} \bibinfo{person}{Ronald Shaffer}.}
  \bibinfo{year}{2010}\natexlab{}.
\newblock \showarticletitle{Facial anthropometric differences among gender,
  ethnicity, and age groups}.
\newblock \bibinfo{journal}{\emph{Annals of occupational hygiene}}
  \bibinfo{volume}{54}, \bibinfo{number}{4} (\bibinfo{year}{2010}),
  \bibinfo{pages}{391--402}.
\newblock


\end{thebibliography}
}

\end{document}